\documentclass[a4paper,conference]{IEEEtran}
\usepackage{amsfonts}
\usepackage{amsmath}
\usepackage{graphicx}
\usepackage{multirow}
\usepackage{array}
\usepackage{picins}
\usepackage{subfigure}
\usepackage{capt-of}
\usepackage{epsfig}
\usepackage{amssymb}
\usepackage{epstopdf}

\ifCLASSINFOpdf
\else
\fi
\begin{document}
\title{Global and Local Consistent Age Generative Adversarial Networks}
\author{\IEEEauthorblockN{Peipei Li,
Yibo Hu,
Qi Li,
Ran He and
Zhenan Sun }
\IEEEauthorblockA{Center for Research on Intelligent Perception and Computing, CASIA, Beijing, China}
\IEEEauthorblockA{National Laboratory of Pattern Recognition, CASIA, Beijing, China}
\IEEEauthorblockA{University of Chinese Academy of Sciences, Beijing, China}
\IEEEauthorblockA{Email: peipei.li, yibo.hu@cripac.ia.ac.cn,  qli,rhe,znsun@nlpr.ia.ac.cn}}
\maketitle

\begin{abstract}
Age progression/regression is a challenging task due to the complicated and non-linear transformation in human aging process. Many researches have shown that both global and local facial features are essential for face representation \cite{su2009hierarchical}, but previous GAN based methods mainly focused on the global feature in age synthesis. To utilize both global and local facial information, we propose a Global and Local Consistent Age Generative Adversarial Network (GLCA-GAN). In our generator, a global network learns the whole facial structure and simulates the aging trend of the whole face, while three crucial facial patches are pregressed or regressed by three local networks aiming at imitating subtle changes of crucial facial subregions. To preserve most of the details in age-attribute-irrelevant areas, our generator learns the residual face. Moreover, we employ an identity preserving loss to better preserve the identity information, as well as age preserving loss to enhance the accuracy of age synthesis. A pixel loss is also adopted to preserve detailed facial information of the input face. Our proposed method is evaluated on three face aging datasets, i.e., CACD dataset, Morph dataset and FG-NET dataset. Experimental results show appealing performance of the proposed method by comparing with the state-of-the-art.
\end{abstract}
% \fi
\IEEEpeerreviewmaketitle
\section{Introduction}
Age progression/regression, also known as age synthesis, aims to aesthetically render given faces with aging or rejuvenating effect but still preserve personality. In recent years, age synthesis has become a hot topic in computer vision. It has a wide range of applications in various domains, e.g., finding missing person, age estimation, age-invariant verification, social entertainment, etc. However, due to the extreme challenges involving diverse genetics and living styles, rigid requirement for training datasets and large variation in illumination, age progression/regression is still a challenging task. In prior works, direct and step-by-step aging synthesis are mainly used for age progression/regression \cite{duong2017learning}. In direct aging synthesis, target age faces can be directly synthesized utilizing the relationships between input faces and their corresponding target age labels. Meanwhile, the step-by-step synthesis usually splits the long-term aging process into short-terms and merely concentrates on the transformation between adjacent age groups. Although the direct synthesis methods are easy to train, they can not synthesize satisfactory results with a long age span and often lose the original identity, while the step-by-step synthesis methods need to train a specific network for every two transformed age groups.

In this paper, the proposed method is based on Conditional Generative Adversarial Nets (CGAN) \cite{mirza2014conditional}, which can simultaneously achieve age progression and regression in the same framework with the given age labels. To remedy identity missing of direct synthesis methods and enhance the accuracy of age synthesis, an identity preserving loss and age preserving loss are also adopted in this method. Moreover, inspired by the observation that for adults, significant aging changes mainly occur in texture of facial subregions and only small changes happen in global facial configurations, we further propose GLCA-GAN with one global network generating the whole facial structure and three local networks processing diverse texture transformation of crucial facial subregions. We also employ a pixel loss to preserve detailed facial information of the input face. In addition, to further preserve most of the details in age-attribute-irrelevant areas, our generator learns the residual face, which is defined as the difference between the input face and its corresponding synthetic face.

The main contributions of this work are as follows:

1) We propose GLCA-GAN for age progression and regression. In GLCA-GAN, one global network is used for generating the whole facial structure and simulating the aging trend of the whole face, while three local networks are employed for processing diverse texture transformation of crucial facial subregions.

2) We impose an age preserving loss to enhance the accuracy of age synthesis. To further ensure that the synthetic face and input face are belong to the same person, an identity preserving loss is adopted. In addition, a pixel loss is also used to preserve detailed facial information.

3) Instead of  manipulating a whole face, our generator learns the residual face between the input face and its corresponding synthetic face, which can accelerate the convergence process while preserving most of the details in age-attribute-irrelevant areas, e.g., clothes, background, etc.

4) Our network can simultaneously achieve age progression and regression in the same framework with the given age labels and generate favorable results.

\section{Related work}
\begin{figure*}[t]
\setlength{\abovecaptionskip}{0cm}
\begin{center}
\includegraphics[width=1\linewidth]{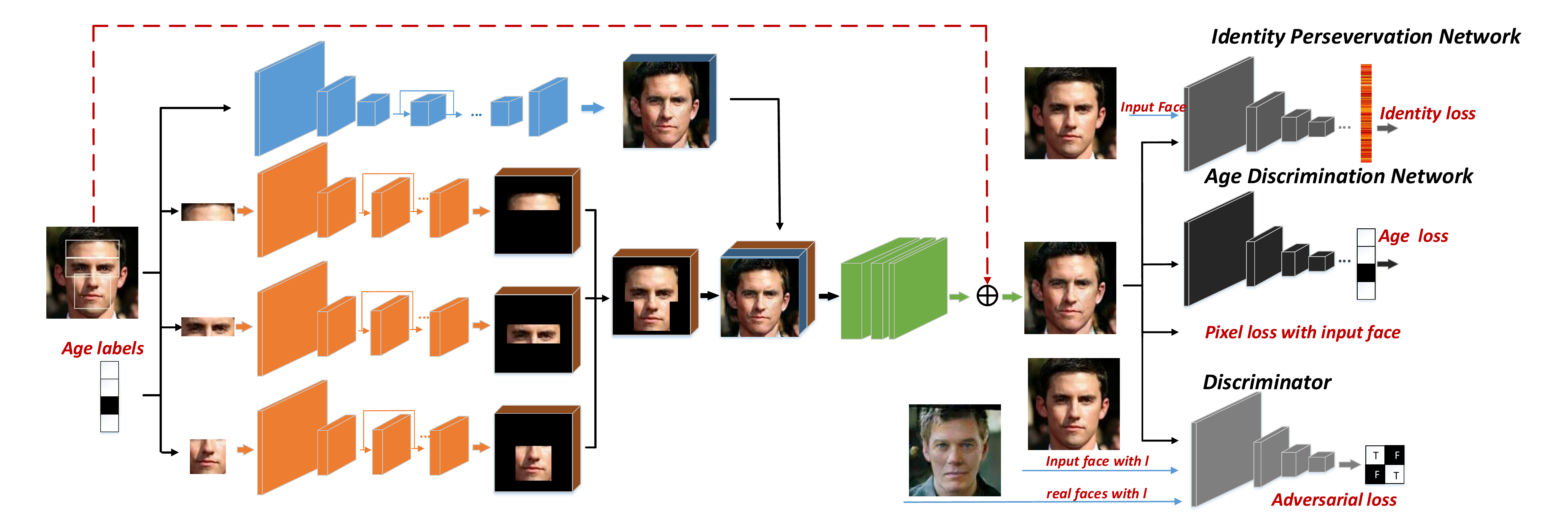}%images//GLA-GAN
\end{center}
\vspace{-0.4cm}
   \caption{General framework of GLCA-GAN. Label $l$ (target age) is concatenated to face image $x$ and three local facial patches $p_1$, $p_2$, $p_3$. The new global tensor $\left[ {x,l} \right]$ is fed to global network, while the three new tensors $\left[{p_1,l} \right]$, $\left[ {p_2,l} \right]$, $\left[ {p_3,l} \right]$ are fed to their corresponding local networks. The outputs of the four networks are fused and fed into three convolutional layers to synthesize the residual face. Besides the discriminator network, an identity network and age discrimination network are also used to strengthen our GLCA-GAN. }
\vspace{-0.5cm}
%\label{fig:synthesis_results}
\end{figure*}
Age progression/regression models can be further divided into physical model-based approaches, prototype-based approaches, and deep learning-based approaches.

The physical model-based approaches \cite{tsai2014human, suo2010compositional, todd1980perception, suo2012concatenational, ramanathan2006modeling}, simulate age progression/regression by modeling human biological and physical mechanisms, e.g., muscles, facial contours, respectively. However, this approach needs many face sequences of the same person covering a long age span, which is hard to obtain.

Meanwhile, prototype-based approaches \cite{tiddeman2001prototyping, kemelmacher2014illumination} regard average faces of each age group as their prototypes, thus the aging pattern is the difference between prototypes of target and source age groups. However, prototype-based approaches may ignore personality, e.g., wrinkles. For this problem, Shu et al.\cite{shu2015personalized} proposed an age synthesis based on aging dictionary learning to reconstruct the aging face utilizing an aging dictionaries of different age groups.

Recently, deep learning-based approaches \cite{nhan2016longitudinal,wang2016recurrent,zhang2017age,yang2017learning,duong2017temporal,zhou2017personalized} have shown considerable results in age synthesis. Wang et al.\cite{wang2016recurrent} proposed a recurrent neural network to make a smoother face aging process. Zhang et al.\cite{zhang2017age} applied Conditional Adversarial Autoencoder (CAAE) to synthesize target age faces with target age labels. In addition, Zhou et al.\cite{zhou2017personalized} argued that occupation information may influence the personal aging process and proposed an occupational-aware adversarial face aging network. To make the most of the image generation ability of GAN, Yang et al.\cite{yang2017learning} put forward a muti-pathway discriminator to refine detailed aging process. Duong et al.\cite{duong2017temporal} presented a generative probabilistic model to simulate the aging process of each age stage.
\section{Method}
In this section, we describe the proposed GLCA-GAN, which contains four parts: a  generator,  a discriminator, an identity preserving network and an age preserving network. The architecture of our method is shown in Fig. 1.
%In our case, we first concatenate target age label $l$ to the input face image $I$ and three local facial patches ${I^{{p_1}}}$, ${I^{{p_2}}}$, ${I^{{p_3}}}$. And then the new global tensor $\left[ {{I^{\rm{g}}},l} \right]$ is fed to global generator to extract structural features, while the three new tensors $\left[ {{I^{{p_1}}},l} \right]$, $\left[ {{I^{{p_2}}},l} \right]$, $\left[ {{I^{{p_3}}},l} \right]$ are fed to their corresponding local generators to encode more detailed variations with marked changed areas. Finally, we fuse the four feature vectors into a new vector and feed it to three convolutional layers to generate synthesis face image. In addition, to shorten the distance between faces with same age in transformed subspace, the input of our discriminator are face images with age labels. Moreover, identity network is employed to push identity features of the input faces and generated faces as close as possible, while age discrimination network is added to distinguish age features in feature level.
\subsection{Generator}
%One global generative network and three local generative networks are adopted in our method to generate accuracy-aging and identity-preserving %images. In addition, every generative network consists of one encoder and one decoder.
Our generator contains a global network and three local networks to learn the whole facial structure and imitate subtle changes of crucial facial subregions simultaneously. To further accelerate the convergence of generator and preserve most of the details of age-attribute-irrelevant areas, our generator learns the residual face between the input face and its corresponding synthetic face. Moreover, the input of our generator is a 128 $\times$ 128 RGB face image ${x} \in {R^{128 \times 128 \times 3}}$ and we concatenate the target age label $l$ to it.

The global network takes the whole image as its input to generate the whole facial structure. First, we utilize three strided convolutional layers to map the input face into a latent space. Then a residual network with four residual blocks acts as a transformer to convert the input embeddings to the target embeddings. After two fractionally-strided convolutional layers, the final output of global network is ${G_g}(x,l) \in {R^{128 \times 128 \times 64}}$.

In addition, we crop three subregions of eyes, snouts, and forehead from the whole face as inputs to three local networks respectively to imitate texture changes of these crucial facial parts. The input local patch $p$ is first processed by two strided convolutional layers, followed by two residual blocks, and after one fractionally-strided convolutional layer, the final output of local network is $G_{l}^i(p,l) \in {R^{128 \times 128 \times 64}},i \in \left\{ {1,2,3} \right\}$.

At last, we fuse the four feature embeddings of global and local networks and feed it to three convolutional layers with stride of 1 to get the residual face. The final output ${G}(x,l) \in {R^{128 \times 128 \times 3}}$ is the adding result of the input and residual faces.

\subsection{Discriminator}

Based on the GAN principle, our discriminator forces the generator to synthesize realistic and plausible faces. Particularly, we impose age labels on discriminator $D$, further forcing the generation of age-specific faces. Both the input face and synthetic face with target age are treated as negative samples, while the real face with target age are as positive samples. To avoid producing artifacts as in \cite{shrivastava2016learning}, our discriminator network distinguishes the $2 \times 2$ local image patches separately and sums the four local cross-entropy losses as the adversarial loss, which can be formulated as:
\begin{equation}
\begin{array}{c}
\;\;{{\rm{L}}_{adv}}\;{\rm{ = }}\;\mathop {\min }\limits_G \mathop {{\rm{max}}}\limits_D {E_{{I_l}{\rm{,}}l{\rm{\sim p(}}{I_l}{\rm{,\;}}l{\rm{)}}}}\left[ {{\rm{log}}D\left( {{I_l}{\rm{,\;}}l} \right)} \right]\vspace{0.2cm}\\
\;{\rm{ +   }}{E_{x{\rm{,}}l{\rm{\sim }}p{\rm{(}}x{\rm{,\;}}l{\rm{)}}}}\left[ {{\rm{log}}\left( {{\rm{1 - }}D\left( {{\rm{x,\;}}l} \right)} \right)} \right]\vspace{0.2cm}\\
\;\;\;\;\;\;\;\;\;\;\;\;{\rm{ +   }}{E_{x{\rm{,}}l{\rm{\sim }}p{\rm{(}}x{\rm{,\;}}l{\rm{)}}}}\left[ {{\rm{log}}\left( {{\rm{1 - }}D\left( {G\left( {x,l} \right),l} \right)} \right)} \right],\vspace{0.1cm}
\end{array}
\end{equation}
where $l$ denotes the one-hot age label and $I_l$ denotes the real face of age group $l$. The data distributions are denoted as ${x{\rm{,}}l{\rm{\sim }}p{\rm{(}}x{\rm{,\;}}l{\rm{)}}}$ and ${{I_l}{\rm{,}}l{\rm{\sim p(}}{I_l}{\rm{,\;}}l{\rm{)}}}$. Parameters of generator and discriminator are trained alternately to optimize the min-max problem.
%Note that our age label is concatenated to both four generators and discriminator to guide the input faces generating age-specific faces, but also discriminator making it discriminate age besides human face.

\subsection{Identity Preserving Loss}

For age progression/regression, it is crucial to keep identity information in the synthesis processes. However, the synthetic faces based on GAN are close to real data only in pixel space, not in semantic space.  Hence, we introduce an identity preserving loss to our model. Furthermore, to preserve finer identity information, we extend the global feature map constraint as \cite{li2018Prot}. Then the identity preserving loss can be formulated as:
\begin{equation}
\begin{array}{c}
{L_{id}} = {\left\| {{\varphi _f}\left( x \right) - {\varphi _f}\left( {G\left( x,l \right)} \right)} \right\|_2}\vspace{0.2cm}\\
\;\;\;\;\;\;\; + \;{\left\| {{\varphi _p}\left( x \right) - {\varphi _p}\left( {G\left( x,l \right)} \right)} \right\|_F},\vspace{0.1cm}
\end{array}
\end{equation}
where ${\varphi _f}$ and ${\varphi _p}$ denote the feature extractors of the fully-connected layer and the last pooling layer of the pre-trained light CNN-29 network \cite{wu2015light} respectively.

\subsection{Age Preserving Loss}
Aging accuracy is another key issue of age progression/regression. In this method, an age preserving loss based on the pre-trained light CNN-29 structure is utilized to enhance the age accuracy of the synthetic face, which can be formulated as:
\begin{equation}
\begin{array}{c}
{L_{age}} =  - \frac{1}{N}\sum\limits_{i = 1}^N {\sum\limits_{j = 0}^M {\left( {1\left\{ {{y^i} = j} \right\}\log \left( s \right)} \right)} },
\end{array}
\end{equation}
where $s$ denotes the output of the final softmax layer of light CNN-29, $y^i$ denotes the class of the i-th training data. And the numbers of training data and age categories are denoted as $N$, $M$ respectively. In addition, when the class of the i-th training data is equal to the j-th class, $1\left\{ {{y^i} = j} \right\} = 1$.

\subsection{Pixel Loss}
To improve the quality of the synthesis face and further preserve the details in age-attribute-irrelevant areas, a pixel-wise $L_2$ loss is adopted, which can be formulated as:
\begin{equation}
\begin{array}{c}
{L_{pixel}} = \frac{1}{{C \times H \times W}}\left\| {x - G\left( x, l \right)} \right\|_2^2,
\end{array}
\end{equation}
where $C$, $H$, $W$ are the channel, height and width size of the tensor face image respectively. For there is no ground-truth data in our experiment, the synthetic face is forced by the pixel-wise loss to  have similar content with the input face. However, it is also essential to make aging or rejuvenating effect on the age-attribute-relevant areas. Thus, we update pixel-level critic at every 5 iteration to balance aging accuracy and identity permanence, which are regarded as the two critical requirements in age progression/regression \cite{yang2017learning}.

\subsection{Objective}
\begin{figure*}[t]
\setlength{\abovecaptionskip}{0cm}
\begin{center}
\includegraphics[width=0.88\linewidth]{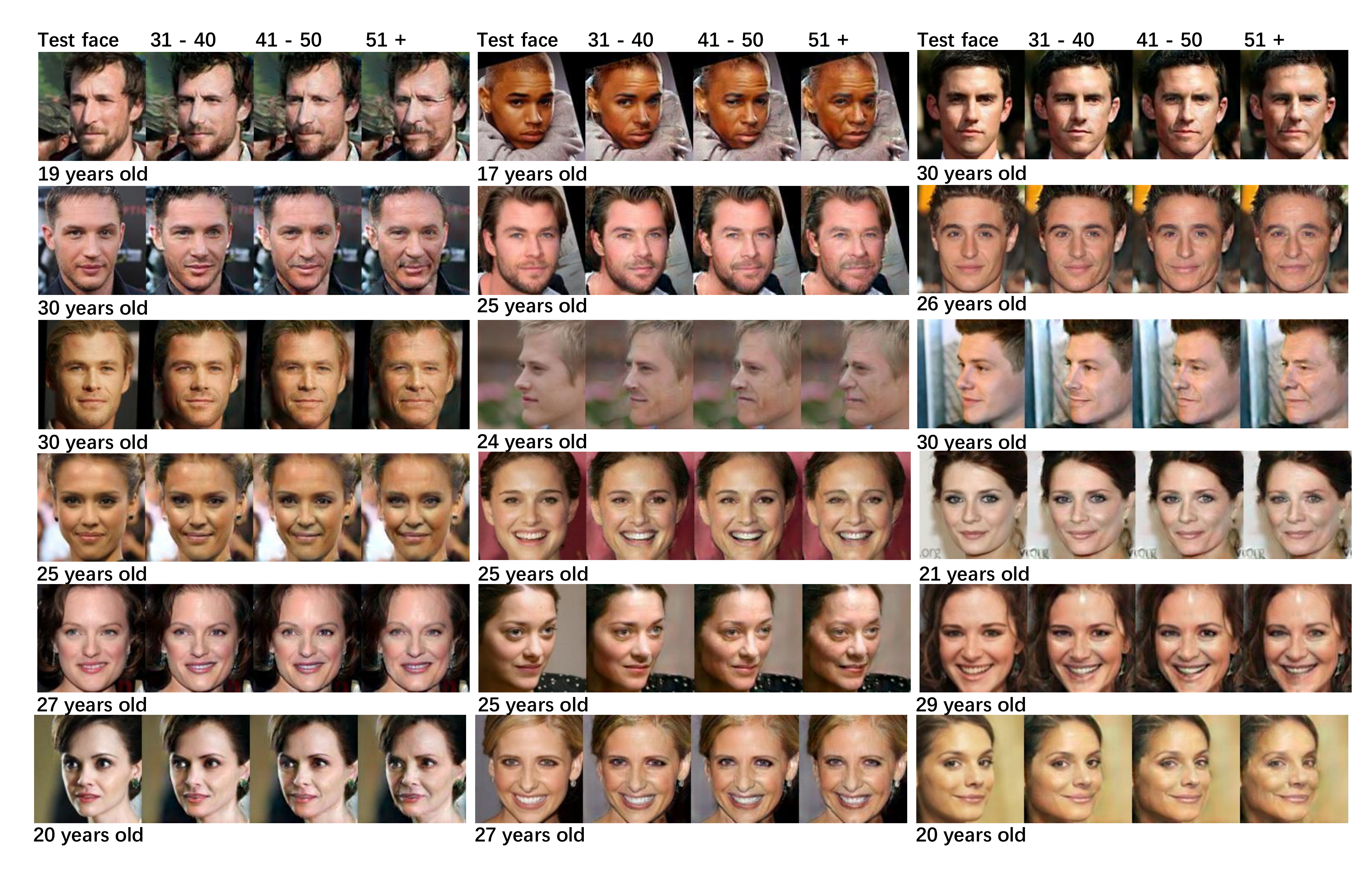}%images//YTF//roc.pdf
\end{center}
\vspace{-0.6cm}
   \caption{Age progression results by GLCA-GAN on CACD dataset for 18 different subjects. For each subject, the leftmost column shows input face, while the rest three columns are synthetic faces from younger to older. (Best viewed in color.)}
\vspace{-0.6cm}
\label{fig:synthesis_results}
\end{figure*}
\begin{figure*}[t]
\setlength{\abovecaptionskip}{0cm}
\begin{center}
\includegraphics[width=0.88\linewidth]{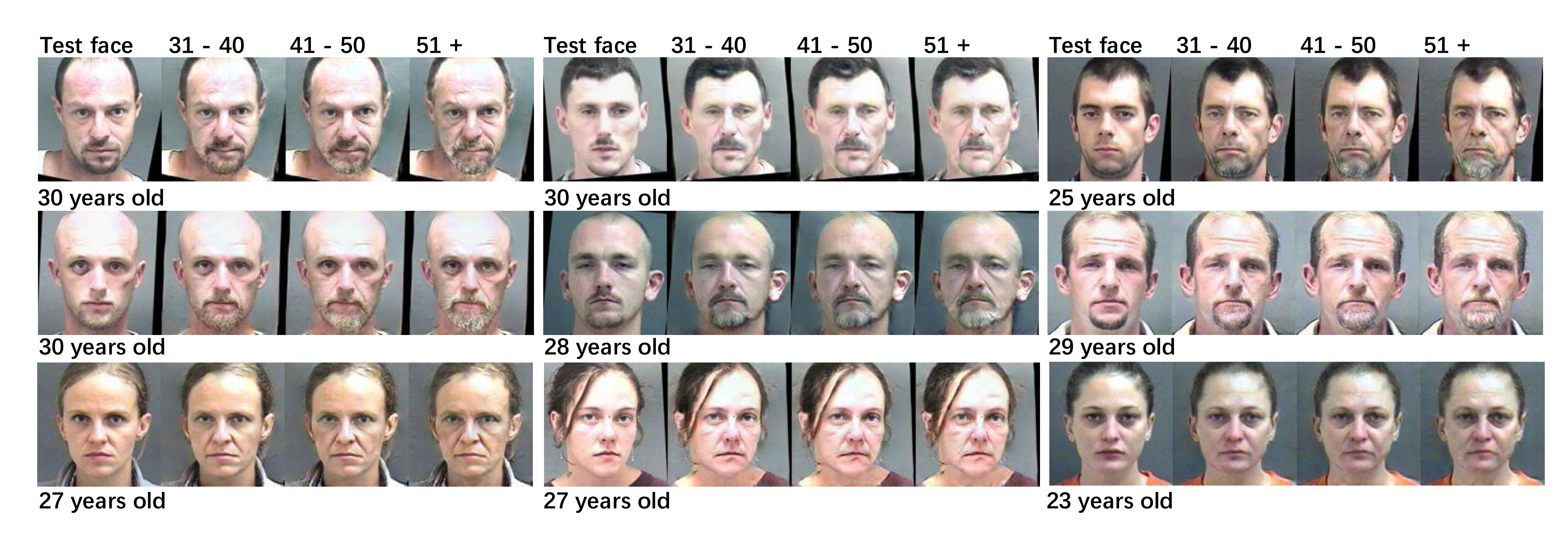}%images//YTF//roc.pdf
\end{center}
\vspace{-0.6cm}
   \caption{Age progression results by GLCA-GAN on Morph dataset for 9 different subjects. For each subject, the leftmost column shows input face, while the rest three columns are synthetic faces from younger to older. (Best viewed in color.)}
\vspace{-0.6cm}
\label{fig:synthesis_results}
\end{figure*}
\begin{figure*}[t]
\setlength{\abovecaptionskip}{0cm}
\begin{center}
\includegraphics[width=0.88\linewidth]{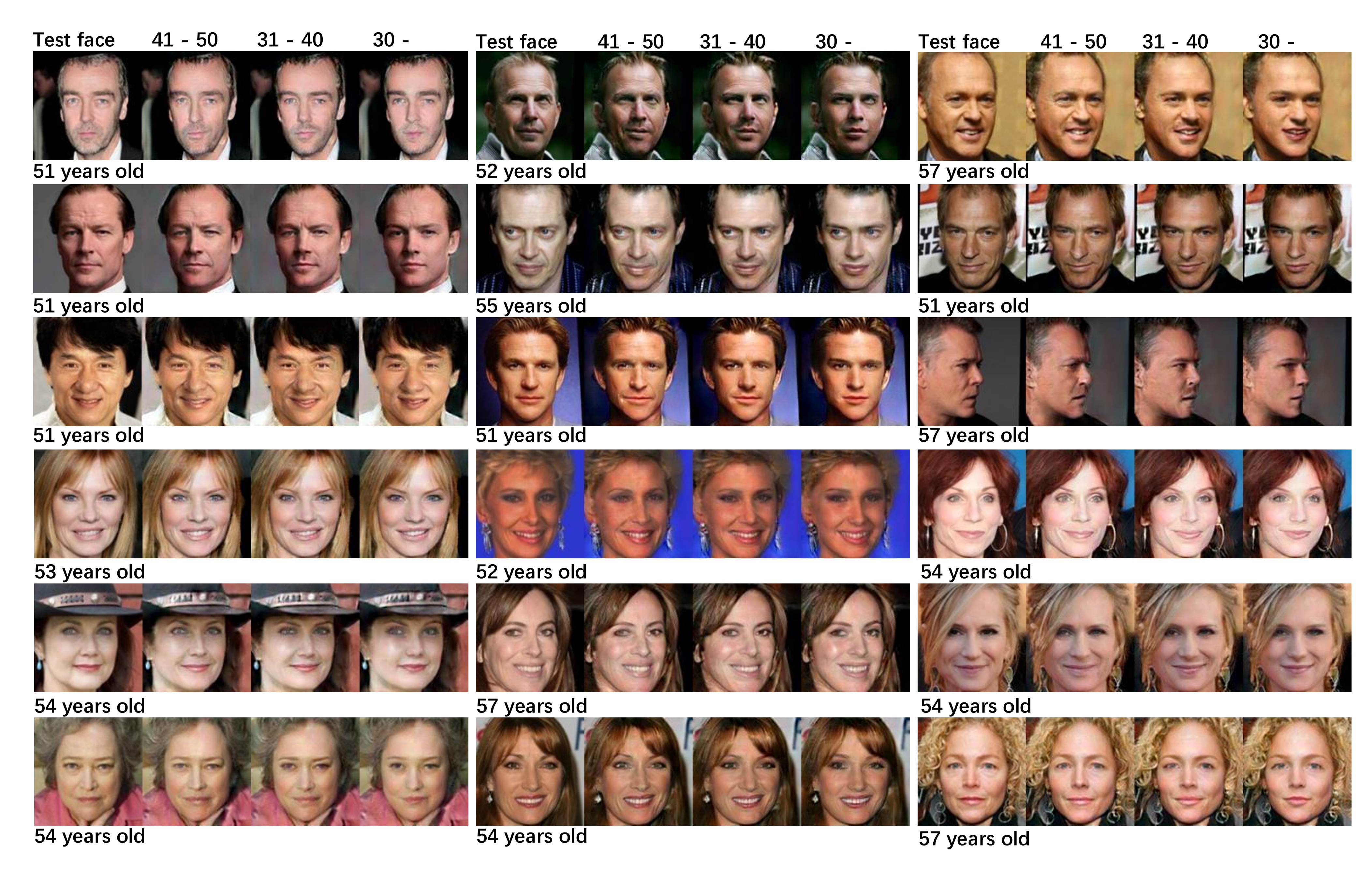}%images//YTF//roc.pdf
\end{center}
\vspace{-0.7cm}
   \caption{Age regression results by GLCA-GAN on CACD dataset for 18 different subjects. For each subject, the leftmost column shows input face, while the rest three columns are synthetic faces from older to younger. (Best viewed in color.)}
\vspace{-0.7cm}
\label{fig:synthesis_results}
\end{figure*}
Taking all loss functions together, the objective function can be expressed as:
\begin{equation}
\begin{array}{c}
{L} = {\lambda _1}{L_{adv}} + {\lambda _2}{L_{ip}} + {\lambda _3}{L_{age}} + {\lambda _4}{L_{pixel}},
\end{array}
\end{equation}
where ${\lambda _1}$, ${\lambda _2}$, ${\lambda _3}$, ${\lambda _4}$ are trade-offs.
%We use ${L_{{\rm{adv}}}}$ to distinguish the real aging face from the synthetic aging face, and ${L_{{\rm{ip}}}}$ to preserve identity information. ${L_{{\rm{age}}}}$ is utilized to achieve accurate aging and ${L_{{\rm{pixel}}}}$ is applied to keep content consistent with the input images.
\section{Experiments}
\subsection{Data Collection}
\begin{figure*}[t]
\setlength{\abovecaptionskip}{0cm}
\begin{center}
\includegraphics[width=0.88\linewidth]{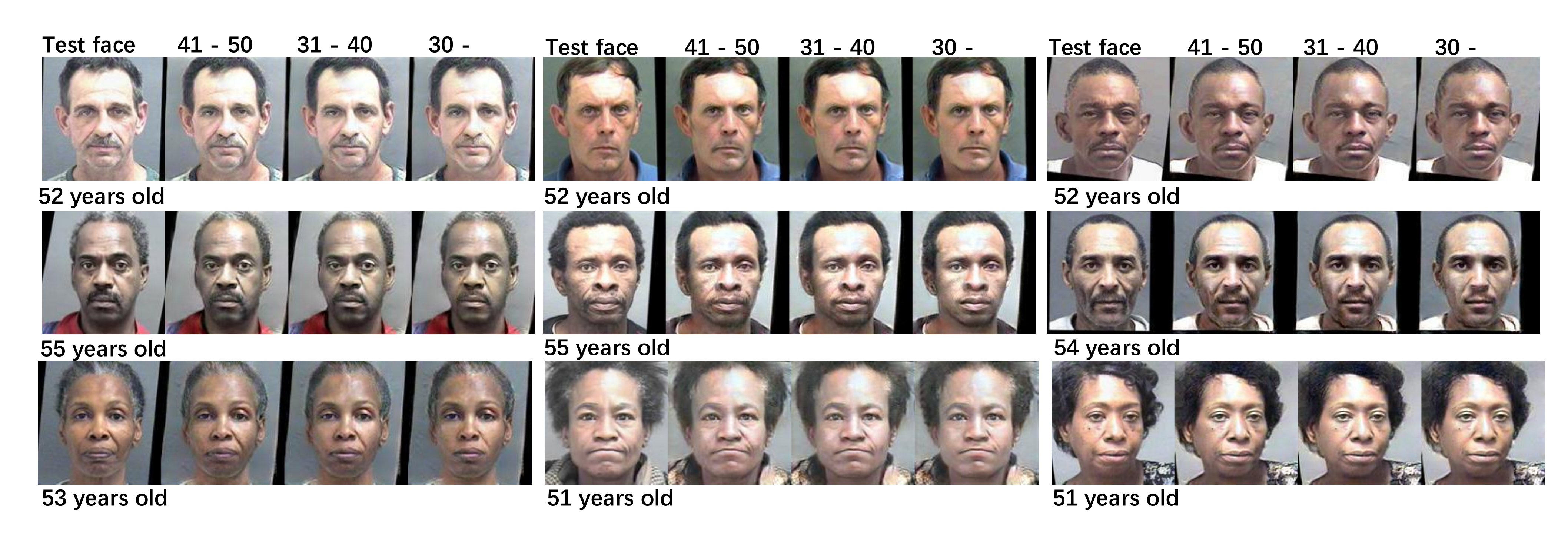}%images//YTF//roc.pdf
\end{center}
\vspace{-0.6cm}
   \caption{Age regression results by GLCA-GAN on Morph dataset for 9 different subjects. For each subject, the leftmost column shows input face, while the rest three columns are synthetic faces from older to younger. (Best viewed in color.)}
\vspace{-0.6cm}
\label{fig:synthesis_results}
\end{figure*}
We collect face images from three available face aging datasets, CACD dataset \cite{chen2015face} , Morph (Album2) dataset \cite{ricanek2006morph} and FG-NET dataset \cite{lanitis2002toward}. Our model is mainly trained and verified on CACD and Morph datasets, while evaluated on FG-NET dataset. The Morph dataset is the largest publicly available longitudinal face database, which contains 55,349 color images of 13,672 subjects with age and gender information. The subject ages of Morph dataset range from 16 to 77 years old. The CACD dataset contains 163,346 color images of 20,000 subjects collected from Internet. The subject age of CACD ranges from 14 to 62 years old. The FG-NET dataset contains 1,002 images of 82 subjects. To train a high-accuracy aging network, it is important to collect exactly labeled face data. As the age information in CACD is not accurate, we manually removed some mismatched images. Finally, 153,106 images of 20,000 subjects are used in our experiment.
\subsection{Implementation Details}
In this experiment, we perform five-fold cross-validation on CACD dataset and Morph dataset respectively. More specifically, we divide each dataset into five folds with one fold being used for testing and the other four folds for training. FG-NET is also adopted as testing set to make comparisons with prior works.

We align the faces of Morph, CACD and FG-NET based on five facial landmarks and crop the aligned faces to $128 \times 128$ pixels. As both CACD and Morph have limited number of faces older than 62 years old or faces younger than 16 years old, we just simulate age progression/regression on faces between 14 to 62 years old. We divide the face data into four age groups: 14-30, 31-40, 41-50, 51-62. During training, we choose Adam optimizer with ${\beta _1}$ of 0.5, ${\beta _2}$ of 0.99 and learning rate of $1 \times {10^{ - 4}}$, and our batch size is set to 10.

For Morph, the trade-off parameters ${\lambda _1}$, ${\lambda _2}$, ${\lambda _3}$ and ${\lambda _4}$ are set to 1.00, 0.005, 20.00, 10.00 respectively; for CACD, they are set to 2.00, 0.01, 25.00, 15.00.
In addition, we update pixel-level critic at every 5 iteration, and update the discriminator for every 2 generator iterations.
%-------------------------------------------------------------------------
\subsection{Performance Comparison}
%In our experiment, we perform five-fold cross-validation on THE CACD dataset and Morph (Album2) dataset respectively. More specifically, we divide each dataset into five folds with one fold being used for testing and the other four folds for training. FG-NET is also adopted as testing sets to make comparisons with prior works.

\subsubsection{Age Progression and Regression}
With an input face and its target age label, GLCA-GAN can directly synthesize target age faces. For age progression, the input faces are under 30 years old. As shown in Fig. 2 and Fig. 3, the identities of all subjects are well preserved, furthermore, the synthetic faces are gradually getting older with deeper nasolabial folds and crow's feet, more white beards and hairs and so on. More importantly, our GLCA-GAN can achieve realistic age progression even on profiles.
Meanwhile, for age regression, the input faces are beyond 51 years old and the results are shown in Fig. 4 and Fig. 5. The GLCA-GAN can turn white beards and hairs into black, vanish wrinkles and so on. During both face aging and rejuvenating, the GLCA-GAN is robust to pose, expression and illumination variations. Detailed facial information, like moles are also well preserved. In addition, changes in age synthesis processes are smooth and consistent.
%\subsubsection{Aging accuracy}
%For aging accuracy evaluation, we employ light CNN-29 as age estimator, which was pretrained by Morph dataset, to judge whether synthetic face belongs to its target age group. As results %shown in Tabel 1, we can see that the verification rates for the three groups are %, %,%,% respectively.
\subsubsection{Identity Preserving}

We evaluate the identity preserving performance of GLCA-GAN on Morph dataset. For age progression, in each fold, about 1,316 faces from different subjects under 30 years old are used as gallery set, while approximately 4,945 synthetic faces are as probe set. For age regression, about 209 faces from different subjects over 51 years old and approximately 669 synthetic faces are used as gallery set and probe set respectively. Light CNN-29 is employed as recognition model in this expriment. As shown in Table 1, the average rank-1 recognition rate of age progression for aged1, aged2, aged3 clusters are 97.66\%, 96.67\%, 91.85\%; meanwhile, the average rank-1 recognition rate of age regression for aged2, aged1, aged0 clusters are 99.64\%, 99.04\%, 98.89\%. As the age gap between the original face and synthetic face increases, the recognition rate decreases, which indirectly proves the accuracy of our age synthesis.

\renewcommand\arraystretch{1.3}

\begin{table}
\setlength{\abovecaptionskip}{0.2cm}
\centering
\caption{Results of face recognition on MORPH(\%).}
\vspace{-0.2cm}
\resizebox{\linewidth}{!}{
%\begin{tabular}{|p{2cm}<{\centering}|p{1cm}<{\centering}|p{1cm}<{\centering}|p{1cm}<{\centering}|p{1cm}<{\centering}|}
\begin{tabular}{c c c c}
\hline
\multirow{1}*{}&\multicolumn{1}{c}{Aged1}&\multicolumn{1}{c}{Aged2}&\multicolumn{1}{c}{Aged3}\\
\cline{2-4}
Progression&97.66 $\pm$ 0.32  &    96.67 $\pm$ 0.58   &     91.85 $\pm$ 2.32\\
\cline{1-4}

&Aged2&Aged1&Aged0\\
\cline{2-4}
Regression&99.64 $\pm$ 0.35  &   99.04 $\pm$ 0.48   &    98.89 $\pm$ 0.60\\
\hline
\end{tabular}}
\vspace{-0.8cm}
\end{table}

\begin{figure}[t]
\setlength{\abovecaptionskip}{0cm}
\begin{center}
\includegraphics[width=0.75\linewidth]{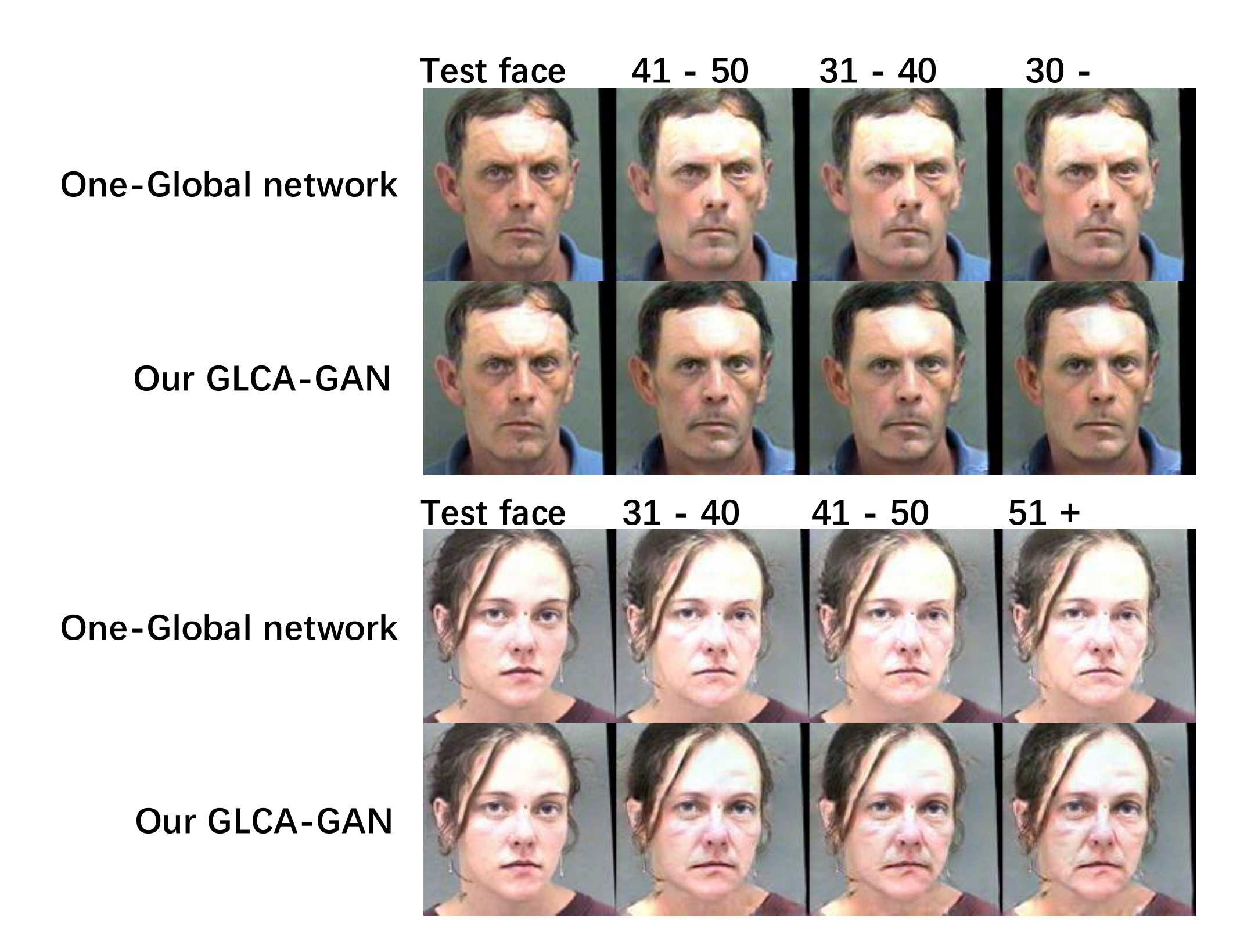}%images//YTF//roc.pdf
\end{center}
\vspace{-0.5cm}
\caption{Comparison to one-global generator on the Morph dataset. (Best viewed in color.)}
\vspace{-0.7cm}
\label{fig:synthesis_results}
\end{figure}

\begin{figure}[t]
\setlength{\abovecaptionskip}{0cm}
\begin{center}
\includegraphics[width=1\linewidth]{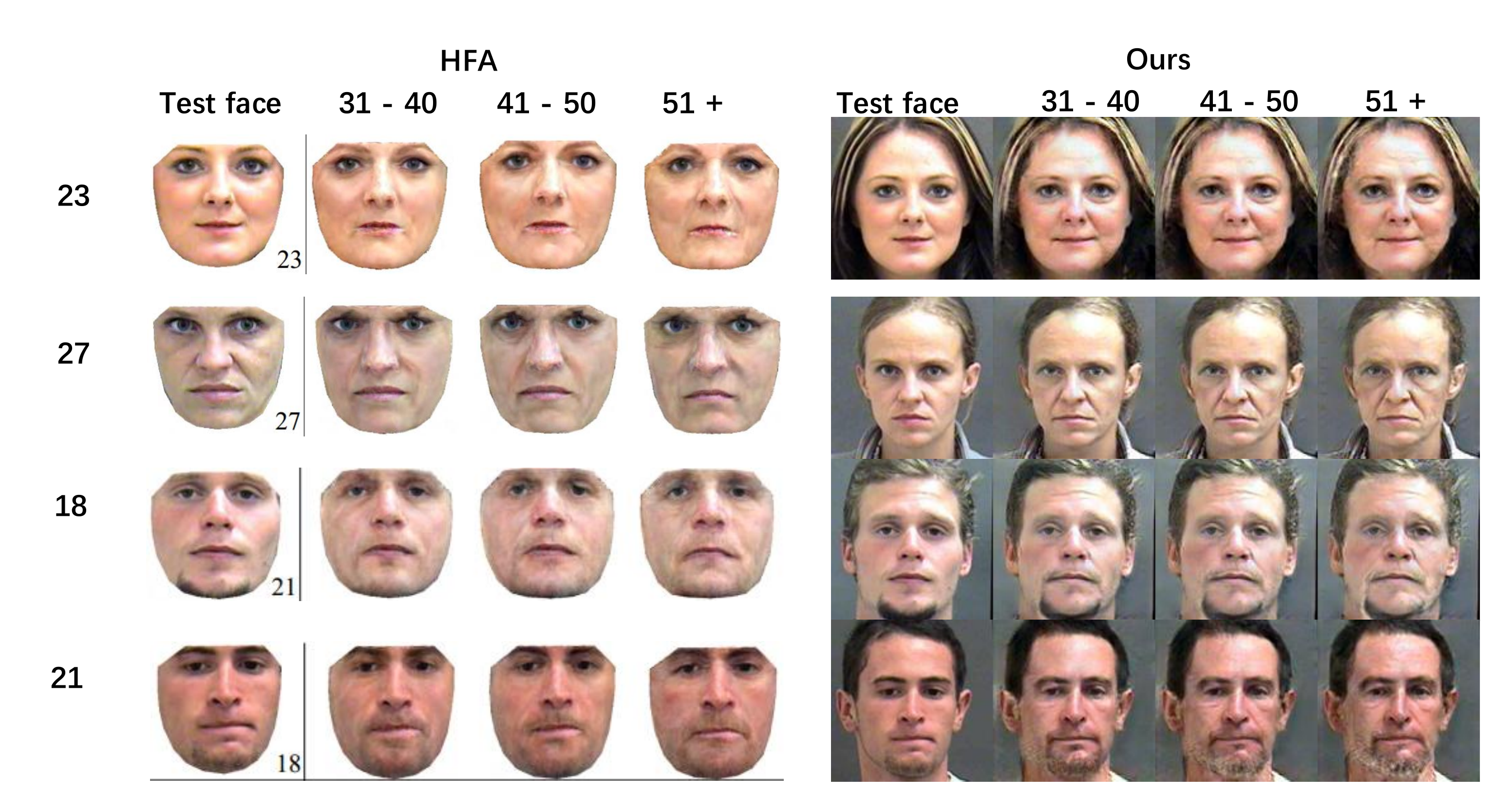}%images//YTF//roc.pdf
\end{center}
\vspace{-0.3cm}
\caption{Comparison to prior works HFA \cite{yang2016face}. (Best viewed in color.)}
\vspace{-0.6cm}
\label{fig:synthesis_results}
\end{figure}

\begin{figure}[t]
\setlength{\abovecaptionskip}{0cm}
\begin{center}
\includegraphics[width=1\linewidth]{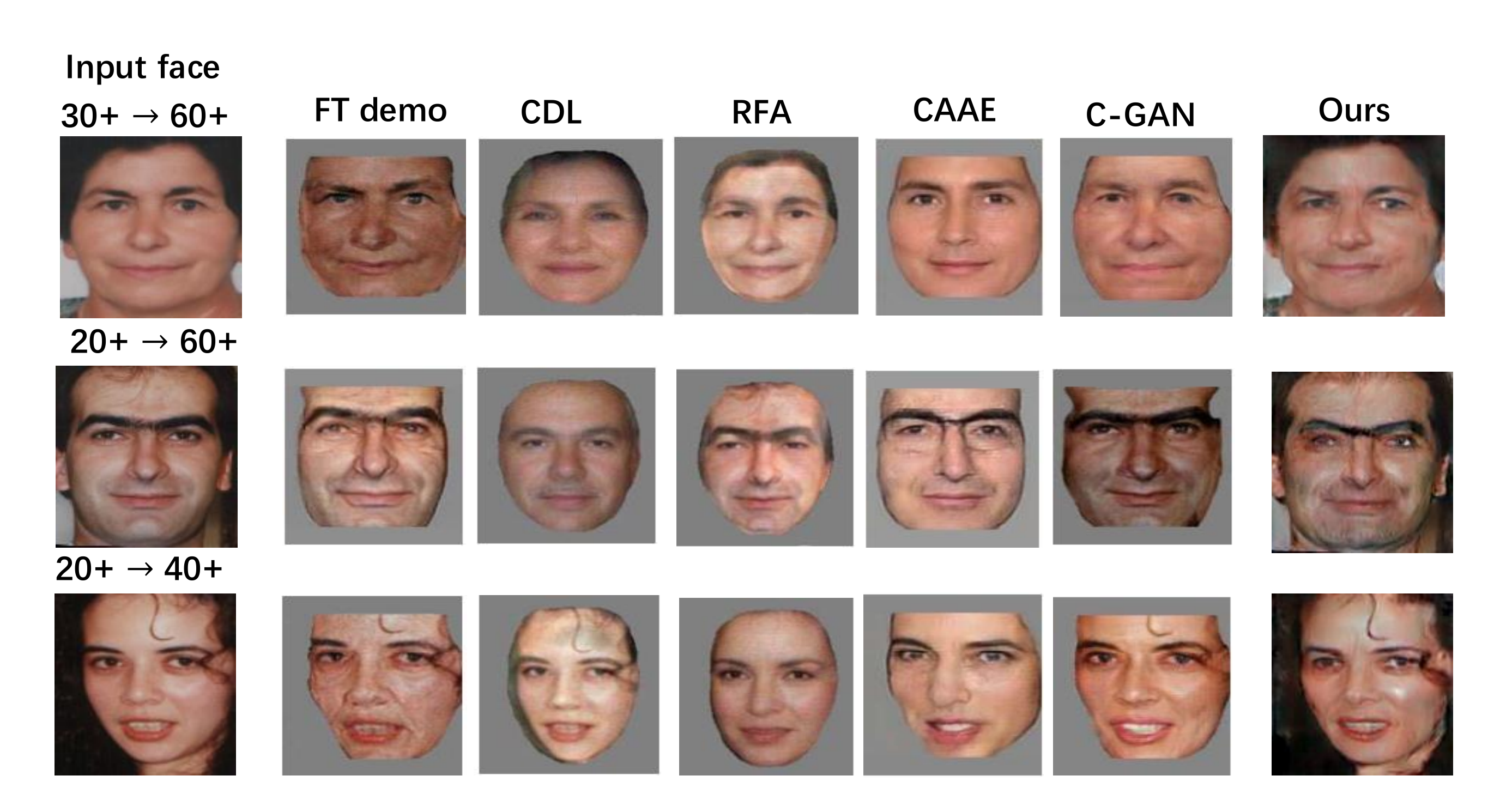}%images//YTF//roc.pdf
\end{center}
\vspace{-0.3cm}
   \caption{Comparison to prior works FT demo \cite{FTdemo}, CDL \cite{shu2015personalized}, RFA \cite{wang2016recurrent}, CAAE \cite{zhang2017age}, C-GAN \cite{liu2017face}. (Best viewed in color.)}
\vspace{-0.6cm}
\label{fig:synthesis_results}
\end{figure}
\subsubsection{Contributions of Local Networks}
To prove the effectiveness of our local networks, we only use global network to synthesize target age faces and keep others consistent with GLCA-GAN. We can see in Fig. 6, both the backgrounds and face contours of input faces are well preserved in the one-global generator and our GLCA-GAN, but our GLCA-GAN works better on synthesizing facial textures, which is probably because of the use of 3 local networks.

\subsubsection{Comparison with Prior Work}
We compare our synthetic results with prior works, including HFA: hidden factor analysis \cite{yang2016face}, FT demo: Face Transformer (FT) demo \cite{FTdemo}, CDL: coupled dictionary learning \cite{shu2015personalized}, RFA: recurrent face
aging \cite{wang2016recurrent}, CAAE: conditional adversarial autoencoder \cite{zhang2017age}, and C-GAN: contextual generative
adversarial nets   \cite{liu2017face}. Furthermore, for fair comparison, we choose the same  faces with their works as our input, and directly cite their synthetic results as most of prior works. We can see from Fig. 7 and Fig. 8 that \cite{yang2016face, zhang2017age, FTdemo, shu2015personalized, wang2016recurrent, liu2017face} only focus on cropped faces. In addition, both texture and crucial region changes of their synthetic faces are not clear. However, in our results, hairs and beards of man become grey along in aging process, while in rejuvenating process, hairs and beards of man turn black. Moreover, our network can simultaneously achieve age progression and regression in the same framework and generate favorable results with background well preserved.

\section{Conclusion}
In this paper, we have proposed a novel Global and Local Consistent Age Generative Adversarial Network (GLCA-GAN) for age progression and regression. Our generator contains one global network and three local networks to learn the whole facial structure and imitate subtle changes of crucial facial subregions simultaneously. To accelerate the convergence and preserve most of the details in age-attribute-irrelevant areas, our generator learns the residual face instead of the whole face. We further introduce an age preserving loss to constraint the synthesis of age-specific faces. Furthermore, an identity preserving loss is imposed to make sure that the input face and synthetic face are of the same person. The pixel loss is also adapted to preserve detailed facial information. Experimental results on CACD, Morph and FG-NET demonstrate the flexibility, generality and efficiency of our method for age progression and regression.


%\section*{Acknowledgment}




\bibliographystyle{IEEEtran}
\bibliography{latex12}
%\begin{thebibliography}{1}


%\end{thebibliography}

\end{document}